\documentclass{article}
\usepackage{amd}
\usepackage[utf8]{inputenc}
\usepackage[T1]{fontenc}
\usepackage{booktabs}
\usepackage{amsfonts}
\usepackage{amsmath}
\usepackage{amssymb}
\usepackage{nicefrac}
\usepackage{microtype}
\usepackage{wrapfig}
\usepackage{tikz}
\usetikzlibrary{arrows.meta, positioning, fit, calc}
\usepackage{placeins}
\usepackage{float}
\usepackage{hyperref}
\hypersetup{hidelinks}
\newcommand{\tmark}{\texttrademark{}}
\IfFileExists{assets/AMD_logo.png}{\newcommand{\AMDlogo}{\includegraphics[height=0.28in]{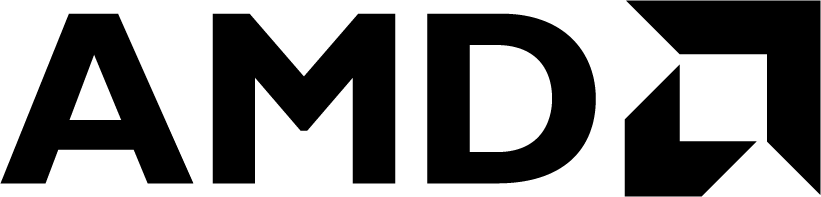}}}{}
\newcommand{\workshopnote}{NeurIPS 2026 Workshop on SLMs for Agentic Systems.}
\makeatletter
\newcommand{\plainfootnote}[1]{\begingroup\let\thefootnote\relax
  \long\def\@makefntext##1{\noindent##1}\footnotetext{#1}\endgroup}
\makeatother

\icmlsetsymbol{intern}{$\dagger$}
\begin{document}
\onecolumn
\amdtitlerunning{Context-Aware Precision Scheduling}
\amdtitle{Context-Aware Precision Scheduling for Small Language Models}

\begin{amdauthorlist}
  \amdauthor{Mukund Athmaram}{amd,cornell,cor,intern}
  \amdauthor{Manoj Rao}{amd}
  \amdauthor{Liang Ma}{amd}
\end{amdauthorlist}
\amdaffiliation{amd}{AMD}
\amdaffiliation{cornell}{Cornell University}
\amdcorrespondingauthor{Mukund Athmaram}{ma2388@cornell.edu}
\amdkeywords{context length, KV cache, llama.cpp, mixed-precision quantization, on-device inference, small language models}
\vskip 0.3in
\amdPrintAffiliationsInline
\plainfootnote{\textsuperscript{$\dagger$}Work conducted during an internship at AMD.\par\noindent\workshopnote}

\begin{abstract}
Quantized small language models (SLMs) running as on-device agents must balance accuracy loss against latency as
their context grows. Mixed precision assigns different precisions to weight tensors according to their size and
sensitivity, but a configuration chosen for a short context may not give the best tradeoff at a longer one. We
therefore change weight precision as context grows. To decide when and which tensors to change, we calibrate a timing
model from measurements at different precisions and context lengths, and combine it with tensor sensitivities to
choose a configuration for each context range under a latency or accuracy budget. Under a latency budget, reading the
growing key--value (KV) cache leaves less time for weight operations, so the allocator lowers precision as context
grows. Under an accuracy budget, the measured error of a quantized KV cache sets the budget left for the weights, so
the allocator can raise or lower weight precision. The result is a schedule of tensor changes prepared before
execution, which our runtime follows without extra weight copies in memory. We compare three ways to do this:
streaming each tensor version from storage, quantizing in place, and storing a low-precision base plus a residual.
On a laptop GPU, we evaluate executed schedules through 16K tokens and on a tool-use benchmark that measures response
time and correctness.
\end{abstract}

\section{Introduction}
Small language models can run on-device as tool-using agents~\citep{slmagents}, where memory and
bandwidth are limited. Each turn adds instructions, tool outputs, and messages, so the context grows and
changes the work required for later responses. Agents also compact their history when it gets too long, so the
context can shrink again mid-session. A fixed precision configuration must serve both short and
long contexts, even though the best configuration may differ between them. We investigate how much is gained by
choosing a configuration for each context length.

Mixed-precision schemes assign different bit-widths to weight tensors because tensors differ in
their sensitivity to quantization and their size. We combine an estimate of each tensor's timing effect
with measurements of its accuracy impact to choose a configuration under a budget. We measure each
tensor's accuracy impact offline by quantizing that tensor alone, keeping all others at reference precision, and
measuring the Kullback--Leibler (KL) divergence between the reference and quantized models' next-token
distributions on WikiText~\citep{merity2017pointer}.
For speed impact, a simple approach is to use bytes as a proxy. This is a natural assumption: for example,
storing weights in four bits rather than eight roughly halves their storage, so it makes sense
to assume that this also makes execution faster.
However, storage alone does not account for the compute and conversion work of each format's
kernels, and Section~\ref{sec:predict} shows a measured case where Q8 is faster than Q4. We therefore fit a
cost model to latencies measured at different precisions and context lengths, evaluate its predictions on new
mixed-precision configurations, and use it to choose precisions as context grows.

Context growth introduces two different scheduling questions. Under a latency budget, reading a
larger KV cache leaves less time for weight operations. We can predict this change, allocate precision
at successive context lengths, and precompute which tensors should change and when. Under an accuracy
budget, we approximate weight and KV-cache errors as additive. Measured cache error determines
the remaining weight-error budget, allowing precision to move in either direction. We test whether
these choices keep measured error within budget and how well the additive model captures the
executed schedule.

We study both questions on an AMD Radeon\tmark{}~780M laptop GPU. The runtime executes tensor changes while
retaining the KV cache, and our measurements include switching time. We extend execution to 16K
contexts with Q8, Q6, and Q4 (8-, 6-, and 4-bit) weights, execute an accuracy-budget schedule with a quantized KV cache,
and evaluate tensor scheduling on tool-use turns. In each experiment we compare our schedule with fixed
configurations, and we include a schedule that estimates latency from stored weight bytes to show how much the
timing model contributes. We then show that schedules follow agent compaction, where precision rises again as the context shortens, and
measure three ways to execute schedules within the memory of a single model copy. The appendix gives calibration details, additional
comparisons, and an evaluation of the additive accuracy estimate.

\noindent\textbf{Contributions.} (i) A calibrated predictor for mixed-precision latency and an allocator
that uses it to schedule tensor precisions as context grows.
(ii) A runtime that switches tensor precision during decoding, and three ways to do so
without keeping extra weight copies in memory. (iii) On-device measurements of longer-context scheduling, the limits of recovering
accuracy through weight precision, and the quality--latency tradeoff in tool use.

\FloatBarrier
\section{Predicting latency of mixed-precision configurations}
\suppressfloats[t]
\label{sec:predict}
Lower precision reduces weight traffic, but computation and conversion also take time. Under the
roofline view~\citep{williams2009roofline,llmviewer,scalingbook}, a matrix multiplication is limited by either
compute or weight loading. In prefill, each weight is reused across many tokens, so compute dominates, and a
lower-bit format can be slower if its conversion costs outweigh the memory saving.

\begin{figure}[H]\centering
\includegraphics[width=\linewidth]{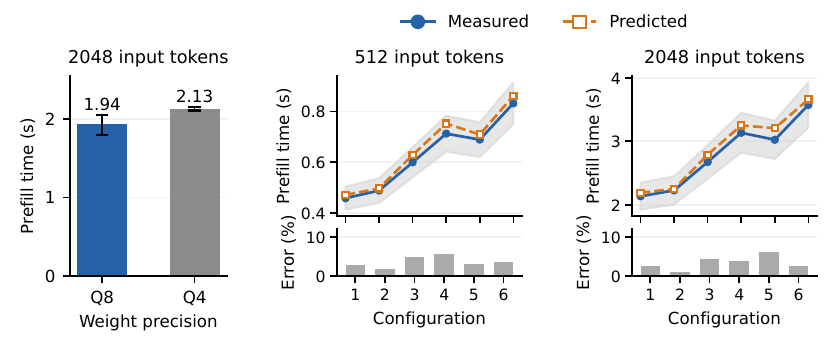}
\caption{Llama-3.2-1B-Instruct~\citep{llama32}, Radeon~780M. Left: Q8 and Q4 prefill latency, with whiskers spanning three passage means. Right: measured (solid) and predicted (dashed) latency for six F16/Q8 configurations excluded from calibration, with $\pm10\%$ bands and percentage error bars.}
\label{fig:latency_predictor}
\end{figure}

We observe this on Llama-3.2-1B-Instruct~\citep{llama32} with native llama.cpp kernels~\citep{llamacpp}
on the Radeon~780M. For a 2048-token prompt, Q8 prefill is faster than Q4 despite storing more weight bytes
(Fig.~\ref{fig:latency_predictor}, left).
Weight bytes therefore do not always determine the runtime ordering.

We contribute a calibrated latency predictor for new mixed-precision configurations. It combines
attention, feed-forward, and output costs using tensors' expected work. Appendix~\ref{app:laptop_calibration}
gives detailed modeling, calibration, and prediction analysis. Section~\ref{sec:calib} specializes
the calibration for context scheduling.

On new mixed-precision configurations, our predictor has lower prefill error than a byte estimate fitted to the same
measurements, and both track decode well (Fig.~\ref{fig:latency_predictor}, right, and
Table~\ref{tab:predictor_errors}). We now use these costs to allocate precision.

\FloatBarrier
\section{Choosing precision at a given context}
\suppressfloats[t]
\label{sec:method}
At context length $L$, we choose a precision $p_u$ for each unit $u$: a weight tensor or a group
that changes precision together. Calibration keeps the device, software, and decoding workload fixed
while context grows.

\FloatBarrier
\subsection{Calibrating context-dependent timing}
\suppressfloats[t]
\label{sec:calib}
\paragraph{Reference latency.} We first measure the reference configuration at several context
lengths and fit
\begin{equation}
T_0(L)=a+c\frac{L}{1024}.
\label{eq:reference_time}
\end{equation}
Here $a$ is the fitted intercept in milliseconds and $c$ is the added time per 1024 context tokens.
This is the total decode-step time, including weight operations, attention, and cache access.
Cache precision stays fixed within each calibration.

\paragraph{Timing changes.} We then time configurations that switch one group of weights (attention, feed-forward, or
embedding/output) to a lower format, and configurations that use that format for every weight. From these runs we
fit one time change $\Delta T_g(p)$ for each group $g$ and format $p$, relative to the reference.
This change includes compute, weight loading, dequantization, and other execution overhead together.
To estimate the time change for an individual tensor, we distribute the group change by weight count:
\begin{equation}
\Delta T(u,p)=\frac{W(u)}{\sum_{v\in g}W(v)}\Delta T_g(p),\qquad u\in g.
\label{eq:timechange}
\end{equation}
Here $W(u)$ is the number of weights in tensor $u$, so each tensor receives a proportional share
of the measured group change. A group allocated as one unit uses $\Delta T_g(p)$ directly.

\paragraph{Configuration latency.} Adding the chosen timing changes to the reference gives
\begin{equation}
T_\text{pred}(\text{cfg},L)=T_0(L)+\sum_u\Delta T\big(u,\text{cfg}[u]\big).
\label{eq:step}
\end{equation}
The fitted quantities are $a$, $c$, and the group changes $\Delta T_g(p)$. Once calibrated,
these coefficients stay fixed as context grows.

\FloatBarrier
\subsection{Allocating precision under a budget}
\suppressfloats[t]
\label{sec:allocator}
We combine timing changes with measured accuracy costs to choose a supported precision
$p_u\in\mathcal{P}_u$. Units with shared costs are allocated together.

\paragraph{Accuracy change.} We use KL to measure prediction changes that aggregate task accuracy
can hide~\citep{accuracy_not_all}. We quantize one unit at a time, keeping the rest at the accuracy
table's reference precision.
Over WikiText~\citep{merity2017pointer}, mean KL between reference and quantized next-token distributions
gives $A(u,p)$. Repeating across units and formats builds the accuracy table. For a change from reference precision $p_0(u)$ to $p$,
\begin{equation}
\Delta A(u,p)=A(u,p)-A\big(u,p_0(u)\big).
\label{eq:daccuracy}
\end{equation}
Positive values mean that changing precision increases divergence.

\paragraph{Choosing the configuration.} For a latency budget $D$, we solve
\begin{equation}
\min_{\{p_u\}}\ \sum_u\Delta A(u,p_u)\quad\text{s.t.}\quad
T_0(L)+\sum_u\Delta T(u,p_u)\le D,\qquad p_u\in\mathcal{P}_u.
\label{eq:knapsack}
\end{equation}
Thus, growing reference latency leaves less room under the same budget. With an accuracy budget
instead, we choose the fastest configuration allowed by the additive accuracy estimate
(Appendix~\ref{app:opt}).
A dynamic program divides the budget into discrete levels and finds the best combination of
precisions on this grid. It is exact for this discretized additive model.
Appendix~\ref{app:opt} illustrates the tradeoff.
Appendix~\ref{sec:opdivergence} compares additive and measured joint KL to evaluate how reliable our additive model
for accuracy impact is.

\FloatBarrier
\section{Executing a context-dependent schedule}
\begin{figure}[H]
\centering
\includegraphics[width=0.8\linewidth]{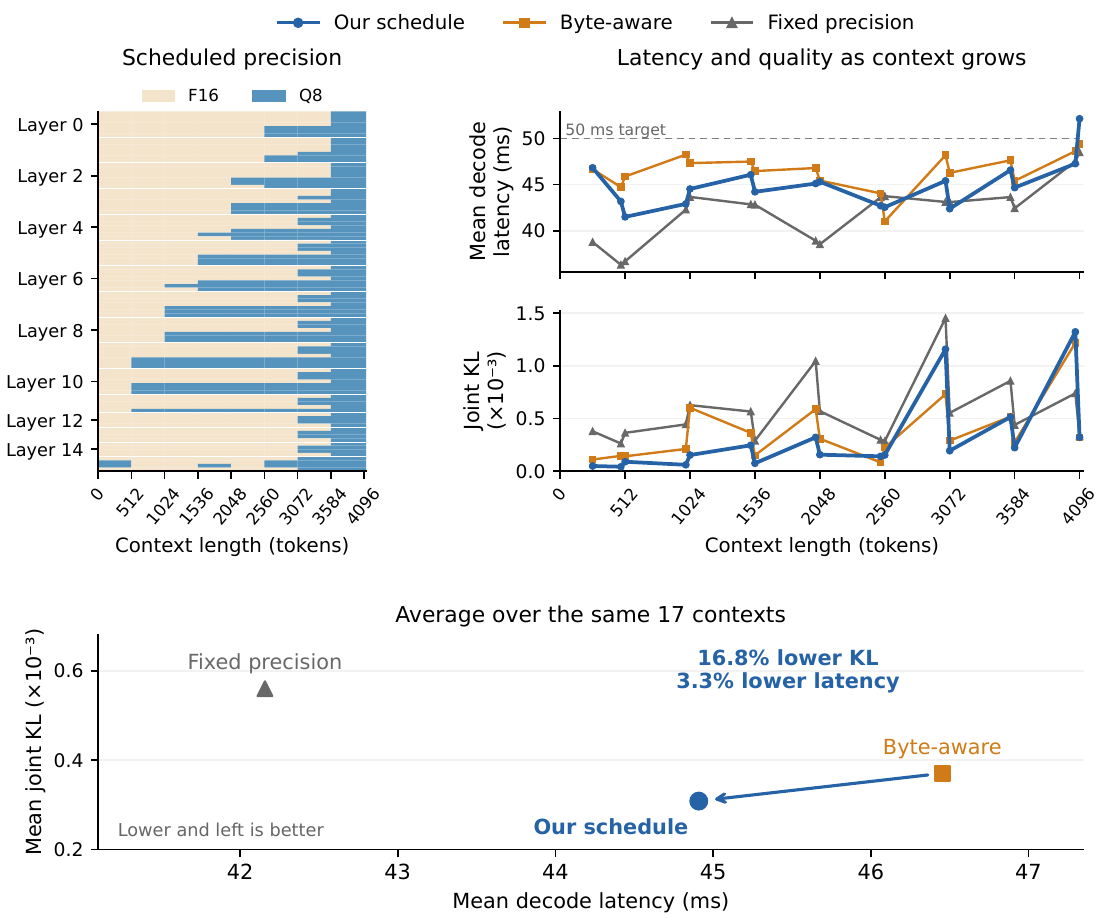}
\caption{Context scheduling on Llama-3.2-1B-Instruct~\citep{llama32}, Radeon~780M. Top: tensor schedule, mean decode latency, and joint KL at matching contexts. Bottom: means over the 17 context lengths. Fixed precision is the F16/Q8 mix selected during calibration.}
\label{fig:ctxsched}
\end{figure}

\suppressfloats[t]
\label{sec:context}
\label{sec:kv}
As context grows, reading the KV cache takes longer, so a fixed latency budget leaves less time for the weights. We
therefore choose a configuration for each range of context lengths, evaluating Eq.~\ref{eq:step} at the longest
context in the range and leaving room for prediction error and switching time. Comparing the configurations of
consecutive ranges gives the schedule, which tensors change and at what context length
(Figure~\ref{fig:ctxsched}, for Llama-3.2-1B-Instruct~\citep{llama32}).

Our main comparison is between the schedule and fixed configurations. To see how much the timing model contributes,
we also include a byte-aware schedule, built with the same allocator and accuracy measurements but with latency
estimated from the bytes used to store the weights.

\paragraph{Executing the schedule.} We modify llama.cpp~\citep{llamacpp} to keep alternate tensor versions in memory and
switch between them at scheduled context lengths. With every version in memory, a switch takes about 2\,ms, around
$4\%$ of a decode step, and is paid only at a transition (Appendix~\ref{app:tensor_runtime}). Section~\ref{sec:memory}
measures the cost of switching without keeping every version in memory.

\paragraph{Latency and quality as context grows.}
We compare our schedule with five fixed configurations on one complete WikiText passage excluded from
calibration~\citep{merity2017pointer}, measuring joint KL to F16 and perplexity (Figure~\ref{fig:ctxsched}). Compared
with using its final configuration throughout, our schedule has about half the joint KL at $8\%$ higher mean
latency, and it meets the 50\,ms target on most steps.

The timing model also contributes here. With it, the schedule has $17\%$ lower joint KL and $3\%$ lower mean latency
than the byte-aware schedule, and it meets the target on more steps (Appendix Figure~\ref{fig:ctxall}).

\subsection{Three precision levels through 16K context}
\label{sec:long_context}
\begin{figure}[H]\centering
\includegraphics[width=0.8\linewidth]{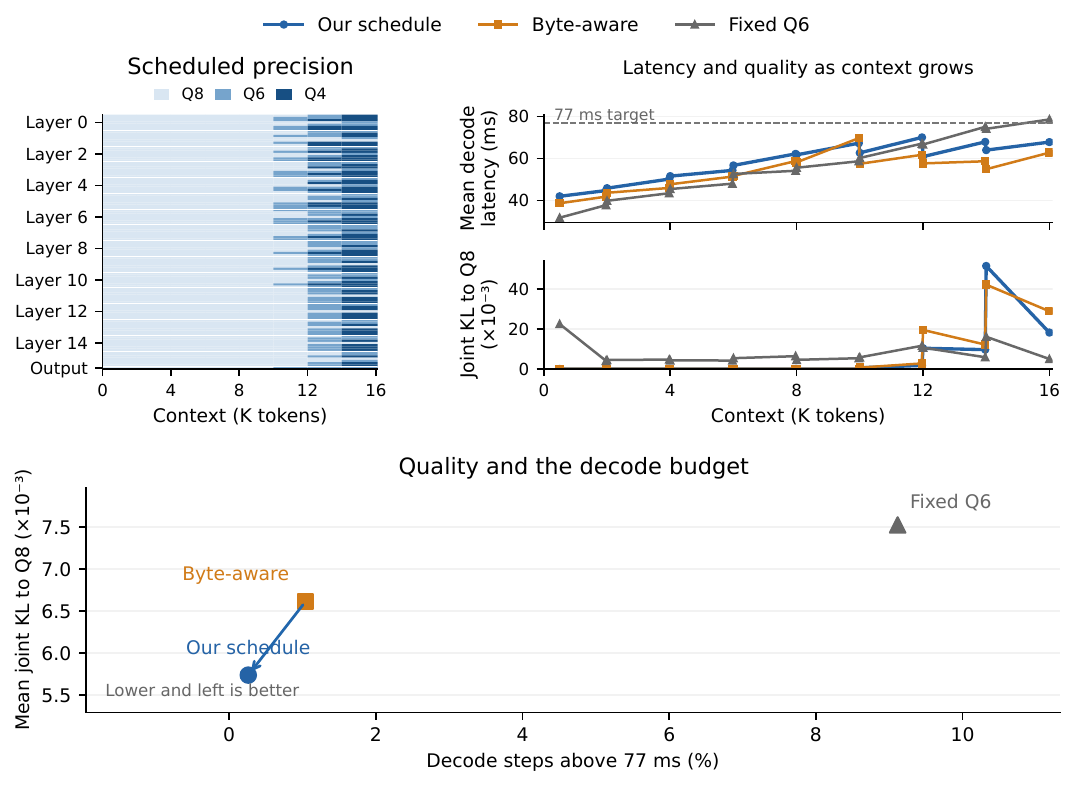}
\caption{Context scheduling through 16K. Top left: the tensor schedule. Right: mean latency and joint KL to Q8 at matching contexts over two evaluation passages, with the 77\,ms target dashed. Bottom: mean KL versus the fraction of decode steps exceeding the target, over the 16 context lengths.}
\label{fig:long_context}
\end{figure}

We extend the runtime to three precision levels, Q8, Q6, and Q4, on Llama-3.2-1B-Instruct~\citep{llama32}, with a
77\,ms target, and evaluate on two WikiText passages excluded from calibration~\citep{merity2017pointer}
(Figure~\ref{fig:long_context}).

Our schedule meets the 77\,ms target on nearly every step, unlike fixed Q8 and Q6, and has lower joint KL than every
fixed configuration except Q8 (Table~\ref{tab:long_results}). Against the byte-aware schedule, which also meets the
target, the timing model gives $13\%$ lower joint KL at $8\%$ higher mean latency.

\FloatBarrier
\section{Scheduling under an accuracy budget}
\suppressfloats[t]
\label{sec:accuracy_context}
\begin{figure}[H]\centering
\includegraphics[width=\linewidth,trim=0 354 0 0,clip]{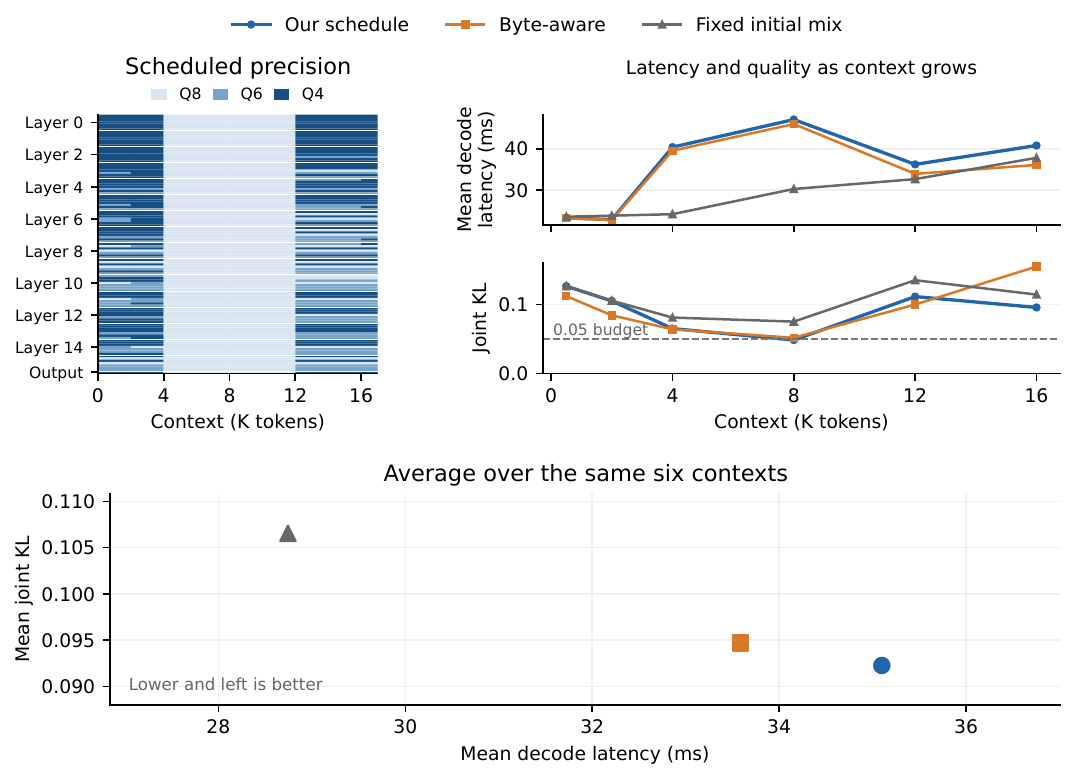}\\[2pt]
\includegraphics[width=0.40\linewidth,trim=0 164 311 20,clip]{figs/slm20261004_accuracy_recovery.pdf}\hfill
\raisebox{2mm}{\includegraphics[width=0.48\linewidth]{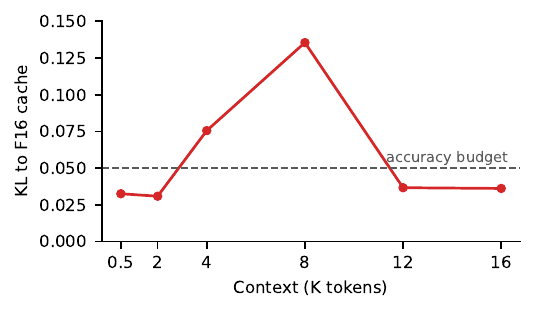}}\\[4pt]
\includegraphics[width=0.603\linewidth,trim=205 164 0 20,clip]{figs/slm20261004_accuracy_recovery.pdf}\\[4pt]
\includegraphics[width=0.85\linewidth,trim=0 0 0 215,clip]{figs/slm20261004_accuracy_recovery.pdf}
\caption{Accuracy-budget scheduling with a Q4 KV cache. Top: tensor precisions chosen by our allocator (left) and
measured cache error at each calibration context, with the accuracy budget dashed (right). Middle: mean decode latency
and joint KL to Q8 weights with an F16 cache at matching contexts, averaged over two passages. Bottom: means over the
six context lengths. Fixed initial mix keeps the first configuration throughout.}
\label{fig:accuracy_recovery}
\end{figure}

We estimate cache error by keeping all weights at Q8 and comparing next-token predictions with
quantized and F16 caches on the same text. At six sampled context lengths, mean KL gives
$A_\mathrm{kv}(\kappa,L)$ for cache precision $\kappa$. We test whether this estimate adds to the
separately measured weight error, leaving an accuracy budget for weights of
\begin{equation}
\epsilon_\mathrm{weights}(L)=\epsilon_\mathrm{total}-A_\mathrm{kv}(\kappa,L).
\label{eq:cachebudget}
\end{equation}
We minimize predicted weight time under this budget (Eq.~\ref{eq:accuracybudget}). Higher cache
error can require higher weight precision.

We evaluate Llama-3.2-1B-Instruct~\citep{llama32} with Q4 keys and values and Q8/Q6/Q4 weights.
We keep the calibrated schedules fixed and evaluate on two WikiText passages~\citep{merity2017pointer}
excluded from calibration, extending context to 16K while retaining each policy's cache. Joint KL is measured against a reference
with Q8 weights and an unquantized F16 cache, and the target is $0.05$ (Appendix~\ref{app:accuracy_recovery}).

The allocator follows the measured cache error (Figure~\ref{fig:accuracy_recovery}). At 4K
and 8K the cache error alone exceeds the budget, so it moves every weight to Q8, and it returns to mixed precision when
the cache error falls again. The cache error does not grow steadily with context. Each point averages the change in next-token predictions at a
few sampled positions, so a few hard-to-predict tokens can dominate it, and at 8K a single token contributes most of the
measured error (Table~\ref{tab:cache_calibration}). Because the evaluation passages differ from the calibration
ones, these spikes do not carry over to them. A schedule should depend on context length rather than on particular
tokens, so averaging over more positions and passages would give a smoother estimate, and calibrating on text from
the target workload would make it more representative. This lowers joint KL relative to keeping the initial configuration, at
higher latency. The byte-aware schedule gives similar joint KL at slightly lower latency, with higher perplexity. The additive estimate of weight plus cache error is optimistic, however, so measured joint KL often
exceeds the budget (Table~\ref{tab:accuracy_results}).

\FloatBarrier
\section{Tool-use evaluation}
\suppressfloats[t]
\label{sec:agents}
We next evaluate whether tensor scheduling helps an agent respond to tool-use requests.
We use Qwen2.5-3B-Instruct~\citep{qwen25three} on six previously unused BFCL multi-turn
tasks~\citep{bfcl}, giving 19 turns per policy. Each policy receives the same preceding messages
and tool outputs, generates its own actions, and is scored by the official simulator. This compares
current-turn correctness from the same conversation context. Response time includes prompt processing,
generation, precision changes, parsing, and tool execution.

\begin{figure}[H]\centering
\includegraphics[width=\linewidth]{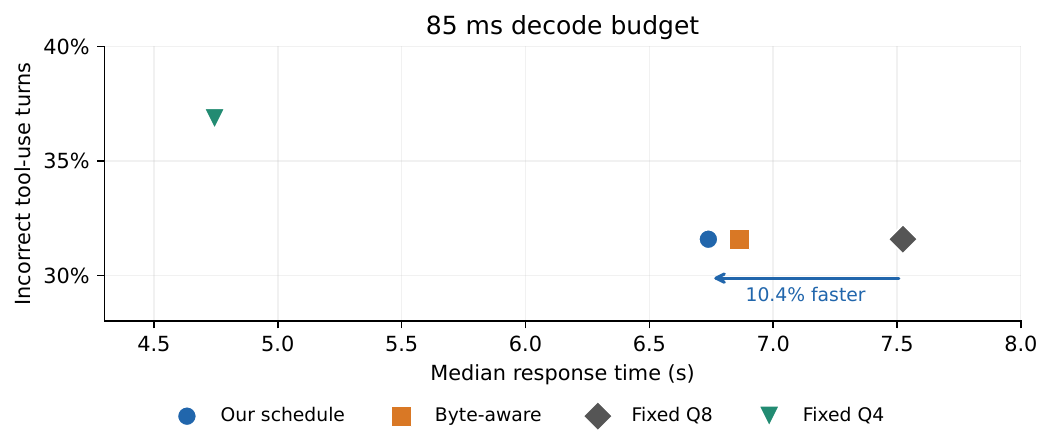}
\caption{Tool-use error versus median complete-turn response time at the 85\,ms decode budget. The arrow marks the response-time difference between our schedule and Q8.}
\label{fig:tensor_agents}
\end{figure}

Our schedule chooses Q8 or Q4 for tensor groups, reusing the KV cache for text shared by successive model calls. We
test 95, 85, and 75\,ms decode budgets against fixed Q8 and Q4, and as in Section~\ref{sec:context} include the
byte-aware schedule to show the timing model's contribution. Figure~\ref{fig:tensor_agents} shows the 85\,ms comparison, and
Appendix~\ref{app:tensor_agents} summarizes all turns and budgets.

Our schedule answers the same 13 of 19 turns correctly as Q8 with $10\%$ lower median response time, and it keeps
nearly every generated token within 85\,ms, while Q8 keeps $60\%$ (Figure~\ref{fig:tensor_agents},
Table~\ref{tab:agent_turns}). Fixed Q4 is faster but answers one fewer turn correctly. The byte-aware schedule is close to ours, so in this
setting the gain comes mainly from scheduling rather than from the timing model.

\FloatBarrier
\section{Agent compaction}
\begin{figure}[H]\centering
\includegraphics[width=\linewidth]{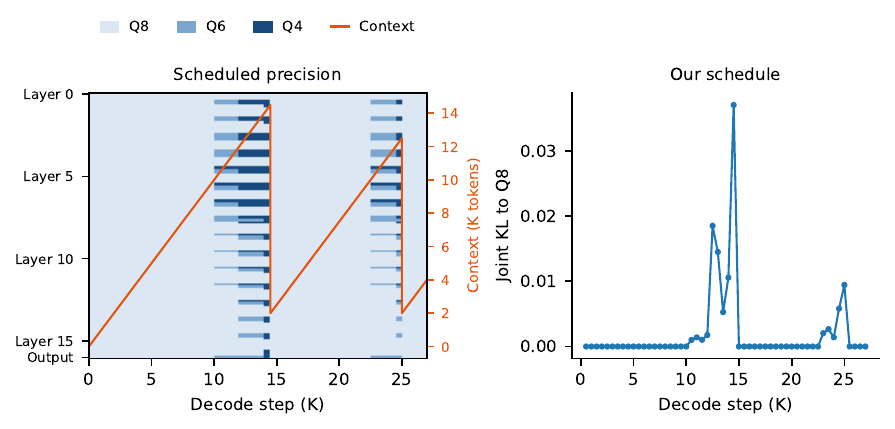}
\caption{A session with two context compactions. Left: scheduled precision of each weight tensor and context
length. Right: joint KL to fixed Q8.}
\label{fig:updown_comp}
\end{figure}

\suppressfloats[t]
\label{sec:compaction}
An agent that compacts its history returns to a short context. Because our schedule depends on context length, not
on how long the session has run, precision returns to Q8 after each compaction, and joint KL falls back toward zero
instead of staying at the level reached before compaction (Figure~\ref{fig:updown_comp}).

\FloatBarrier
\FloatBarrier
\section{How do we switch precision?}
\suppressfloats[t]
\label{sec:memory}
The runtime in Section~\ref{sec:context} keeps every format of every switchable tensor in memory, so a switch only
changes which copy a tensor uses. This more than doubles weight memory compared with the Q8 model alone
(Table~\ref{tab:memory}). Since a schedule is fixed before decoding, the runtime knows which tensor versions it will
need long before it needs them, so it can also run without resident copies. We compare two ways to do this, including for
schedules that raise precision again (Sections~\ref{sec:accuracy_context} and~\ref{sec:compaction}): loading each
tensor version shortly before it is needed, or storing each tensor once in a form that can switch precision itself
(Appendix~\ref{app:memory}).

\begin{figure}[H]\centering
\includegraphics[width=\linewidth]{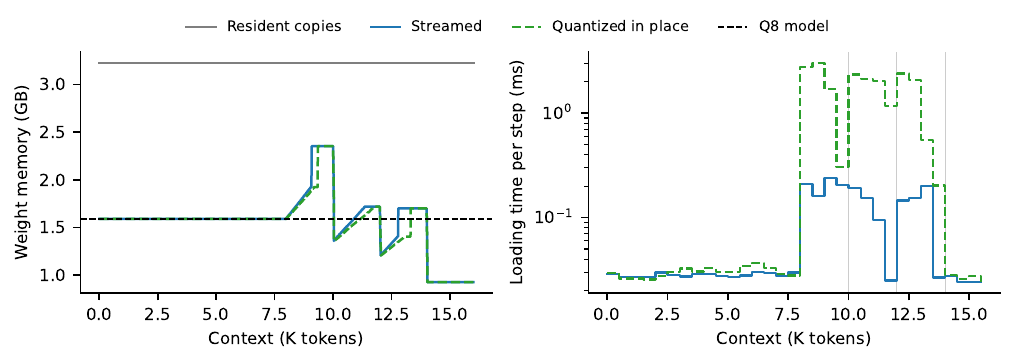}
\caption{Executing the 16K schedule without resident copies (Llama-3.2-1B-Instruct, Radeon~780M). Left:
allocated weight memory as context grows. Right: decode-thread time spent loading, switching, and freeing per step;
gray lines mark the switches.}
\label{fig:stream_memory}
\end{figure}

\paragraph{Loading tensor versions when needed.} Streaming reads the next interval's tensor versions from disk in the
background, switches at the scheduled position, and frees versions no later interval uses. Quantizing in place uses
the same pipeline but quantizes each incoming tensor on the CPU from the most precise version still in memory. Both
trade weight memory against loading work (Figure~\ref{fig:stream_memory}, Table~\ref{tab:memory}).

\begin{table}[H]\centering\small
\caption{Executing the 16K schedule with and without resident copies (two WikiText passages, 16{,}384 decode
steps each).}
\label{tab:memory}
\setlength{\tabcolsep}{4pt}
\resizebox{\linewidth}{!}{\begin{tabular}{lrrrrl}\toprule
Execution & Peak weights & At 16K & Loading time & Waits at switches & Perplexity\\\midrule
Resident copies & 3.23\,GB & 3.23\,GB & -- & -- & 12.542\\
Streamed & 2.35\,GB & 0.93\,GB & 1.1--1.3\,s & none & 12.542\\
Quantized in place & 2.35\,GB & 0.93\,GB & 10.4--11.4\,s & none & 12.566\\
\bottomrule\end{tabular}}
\end{table}

Streaming gives predictions identical to resident copies, has every tensor ready before its switch, and adds little
decode-thread time (Figure~\ref{fig:stream_memory}, Table~\ref{tab:memory}). Its weight memory peaks at the Q8 model
plus one interval of incoming tensors and falls as precision drops. It also follows the schedules that raise
precision again (Tables~\ref{tab:compaction} and~\ref{tab:updown}). Quantizing in place also works but is CPU-bound,
and it cannot follow schedules that raise precision, since it cannot recover a precision it has discarded.

\paragraph{Storing one switchable copy.} Residual branches store each tensor once as a 4-bit base plus a 4-bit
residual, in the spirit of nested and residual formats~\citep{anyprecision,matquant,nestedfp,rrq}, and switching
turns the residual on or off. We add a fused llama.cpp kernel that computes both halves in one pass. Nothing is
read from disk during decoding, and weight memory stays constant at about the size of the Q8 model, so the comparison here is
on speed and accuracy. With the residual on, decode and prefill are slower than Q8 (Table~\ref{tab:residual}). Under a latency budget, this makes the schedule turn off more residuals, so in the
compaction session they have higher joint KL than resident copies at equal latency (Tables~\ref{tab:compaction}
and~\ref{tab:eqlat}). Under the accuracy budget, the residual schedules are slightly more accurate
(Table~\ref{tab:updown}). Streaming is therefore the practical way to execute schedules with unchanged kernels, while
residual branches trade some accuracy at a given latency for constant memory and no loading.

\begin{table}[H]\centering\small
\caption{Compaction session executed three ways (residual branches use a Q4\_K base). Joint KL is the mean at sampled steps over two passages, and loading
is the decode-thread time spent loading and switching tensors.}
\label{tab:compaction}
\begin{tabular}{lrrrr}\toprule
Execution & Peak & Minimum & Loading & Joint KL\\\midrule
Resident copies & 3.23\,GB & 3.23\,GB & -- & 0.0024\\
Streamed & 2.43\,GB & 1.16\,GB & 1.6--1.8\,s & 0.0024\\
Residual branches & 1.67\,GB & 1.67\,GB & 0.7--0.8\,s & 0.010\\
\bottomrule\end{tabular}\end{table}

\begin{table}[H]\centering\small
\caption{Accuracy-budget schedule executed three ways (residual branches use a Q4\_K base). Joint KL is the mean at sampled steps over two passages, the last
column counts samples within the 0.05 budget, and loading is as in Table~\ref{tab:compaction}.}
\label{tab:updown}
\begin{tabular}{lrrrrr}\toprule
Execution & Peak & Minimum & Loading & Joint KL & Within budget\\\midrule
Resident copies & 3.04\,GB & 3.04\,GB & -- & 0.057 & 36/66\\
Streamed & 2.55\,GB & 1.25\,GB & 1.4--1.6\,s & 0.057 & 36/66\\
Residual branches & 1.67\,GB & 1.67\,GB & 0.2--0.3\,s & 0.049 & 45/66\\
\bottomrule\end{tabular}\end{table}

\FloatBarrier
\section{Related work and discussion}
\suppressfloats[t]
\noindent\textbf{Mixed-precision allocation.} Sensitivity and hardware costs guide precision choices in
HAWQ/HAQ~\citep{hawq,haq}, dMX/MxMoE~\citep{dmx,mxmoe}, and LLM-PQ~\citep{llmpq}. \citet{kl_lens}
use forward-only KL sensitivity in hybrid state-space/Transformer models.

\noindent\textbf{Precision scheduling.} PMPD~\citep{pmpd} reduces average bit-width under a quality
constraint with static and prompt-adaptive decoding schedules. MorphServe~\citep{morphserve}
swaps quantized layers as serving load changes, preserving attention state.
QAQ~\citep{qaq_query} learns to choose precision for each query and loads the required weights on demand.
NestedFP~\citep{nestedfp} serves FP16 and FP8 from one stored copy, and Any-Precision LLM~\citep{anyprecision},
Matryoshka Quantization~\citep{matquant}, and BitStack~\citep{bitstack} store weights so that several bit-widths can be
read from one copy, and Recurrent Residual Quantization~\citep{rrq} adds quantized residuals to a low-bit base.
These representations make a switch cheap, and our allocator decides which tensors to switch and when, from calibrated
time and context length. FlexQuant~\citep{flexquant} and DP-LLM~\citep{dpllm} also change layer precision during
decoding, choosing from accuracy signals rather than a latency budget.
We precompute which tensors to change and when under a context-dependent time budget.

\noindent\textbf{KV compression.} KIVI~\citep{kivi} and KVQuant~\citep{kvquant} compress the cache
itself. We use measured cache error to set the weight-error budget. Raising weight precision can
improve predictions but leaves stored cache entries unchanged (Section~\ref{sec:accuracy_context}).

\noindent\textbf{Scope and future work.} We test one laptop GPU, and using a different device or software stack
requires new timing calibration.
Agent turns start from the same conversation context, and our compaction session uses a fixed threshold. Real
agents compact their history with their own policies, and since our model predicts how latency grows with context,
it could also help decide when to compact, trading earlier compaction against lower weight precision.

\FloatBarrier
\section{Conclusion}
\suppressfloats[t]
Our calibrated predictor estimates the latency of mixed-precision configurations and captures
execution costs that weight bytes alone can miss. It provides a way to compare precision choices
before running each configuration.
Combining context-dependent timing with tensor sensitivities gives a precision schedule that can
be prepared in advance and executed during decoding. Our experiments show improved
latency--fidelity tradeoffs and faster tool-use responses at matching correctness, alongside the
limits of recovering cache error through weight precision. We hope serving systems can use
the predictor and scheduling approach to choose precision as a session grows.

\FloatBarrier

\section*{Acknowledgments}
AMD, the AMD Arrow logo, Radeon, and combinations thereof are trademarks of Advanced Micro Devices, Inc. Other product
names used in this publication are for identification purposes only and may be trademarks of their respective
companies.

\bibliographystyle{mlsys2025}
\bibliography{refs}

@misc{dmx,
  title         = {{dMX}: Differentiable Mixed-Precision Assignment for Low-Precision Floating-Point Formats},
  author        = {Franco, Giuseppe and Colbert, Ian and Monteagudo-Lago, Pablo and Marty, Felix and Fraser, Nicholas},
  year          = {2026},
  eprint        = {2606.04115},
  url = {https://arxiv.org/abs/2606.04115},
  archivePrefix = {arXiv},
  primaryClass  = {cs.LG},
}

@inproceedings{hawq,
  title     = {{HAWQ}: Hessian AWare Quantization of Neural Networks with Mixed-Precision},
  author    = {Dong, Zhen and Yao, Zhewei and Gholami, Amir and Mahoney, Michael W. and Keutzer, Kurt},
  booktitle = {Proceedings of the IEEE/CVF International Conference on Computer Vision (ICCV)},
  year      = {2019},
  note      = {arXiv:1905.03696},
}

@inproceedings{haq,
  title     = {{HAQ}: Hardware-Aware Automated Quantization with Mixed Precision},
  author    = {Wang, Kuan and Liu, Zhijian and Lin, Yujun and Lin, Ji and Han, Song},
  booktitle = {Proceedings of the IEEE/CVF Conference on Computer Vision and Pattern Recognition (CVPR)},
  year      = {2019},
  note      = {arXiv:1811.08886},
}

@inproceedings{mxmoe,
  title     = {{MxMoE}: Mixed-Precision Quantization for {MoE} with Accuracy and Performance Co-Design},
  author    = {Duanmu, Haojie and Li, Xiuhong and Yuan, Zhihang and Zheng, Size and Duan, Jiangfei and Zhang, Xingcheng and Lin, Dahua},
  booktitle = {Proceedings of the 42nd International Conference on Machine Learning (ICML)},
  year      = {2025},
  note      = {arXiv:2505.05799},
}

@inproceedings{pmpd,
  title     = {Progressive Mixed-Precision Decoding for Efficient {LLM} Inference},
  author    = {Chen, Hao Mark and Tan, Fuwen and Kouris, Alexandros and Lee, Royson and Fan, Hongxiang and Venieris, Stylianos I.},
  booktitle = {International Conference on Learning Representations (ICLR)},
  year      = {2025},
  note      = {arXiv:2410.13461},
}

@inproceedings{llmpq,
  title     = {{LLM-PQ}: Serving {LLM} on Heterogeneous Clusters with Phase-Aware Partition and Adaptive Quantization},
  author    = {Zhao, Juntao and Wan, Borui and Peng, Yanghua and Lin, Haibin and Wu, Chuan},
  booktitle = {Proceedings of the 29th ACM SIGPLAN Annual Symposium on Principles and Practice of Parallel Programming (PPoPP), Poster},
  year      = {2024},
  note      = {arXiv:2403.01136},
}

@misc{llmviewer,
  title         = {{LLM} Inference Unveiled: Survey and Roofline Model Insights},
  author        = {Yuan, Zhihang and Shang, Yuzhang and Zhou, Yang and Dong, Zhen and Zhou, Zhe and Xue, Chenhao and Wu, Bingzhe and Li, Zhikai and Gu, Qingyi and Lee, Yong Jae and Yan, Yan and Chen, Beidi and Sun, Guangyu and Keutzer, Kurt},
  year          = {2024},
  eprint        = {2402.16363},
  url = {https://arxiv.org/abs/2402.16363},
  archivePrefix = {arXiv},
  primaryClass  = {cs.CL},
}

@misc{scalingbook,
  title         = {How to Scale Your Model: A Systems View of {LLM}s},
  author        = {Austin, Jacob and Douglas, Sholto and Frostig, Roy and Levskaya, Anselm and Chen, Charlie and Vikram, Sharad and Lebron, Federico and Choy, Peter and Ramasesh, Vinay and Webson, Albert and Pope, Reiner},
  year          = {2025},
  howpublished  = {\url{https://jax-ml.github.io/scaling-book/}},
  note          = {Google DeepMind},
}

@inproceedings{kivi,
  title = {{KIVI}: A Tuning-Free Asymmetric 2bit Quantization for {KV} Cache},
  author = {Liu, Zirui and Yuan, Jiayi and Jin, Hongye and Zhong, Shaochen and Xu, Zhaozhuo and Braverman, Vladimir and Chen, Beidi and Hu, Xia},
  booktitle = {Proceedings of the 41st International Conference on Machine Learning (ICML)},
  year = {2024},
  note = {arXiv:2402.02750},
}

@misc{kvquant,
  title = {{KVQuant}: Towards 10 Million Context Length {LLM} Inference with {KV} Cache Quantization},
  author = {Coleman Hooper and Sehoon Kim and Hiva Mohammadzadeh and Michael W. Mahoney and Yakun Sophia Shao and Kurt Keutzer and Amir Gholami},
  year = {2025},
  eprint = {2401.18079},
  archivePrefix = {arXiv},
  primaryClass = {cs.LG},
  url = {https://arxiv.org/abs/2401.18079},
}

@misc{slmagents,
  title         = {Small Language Models are the Future of Agentic {AI}},
  author        = {Belcak, Peter and Heinrich, Greg and Diao, Shizhe and Fu, Yonggan and Dong, Xin and Muralidharan, Saurav and Lin, Yingyan Celine and Molchanov, Pavlo},
  year          = {2025},
  eprint        = {2506.02153},
  url = {https://arxiv.org/abs/2506.02153},
  archivePrefix = {arXiv},
  primaryClass  = {cs.AI},
}

@misc{llamacpp,
  title        = {llama.cpp: {LLM} inference in {C/C++}},
  author       = {Gerganov, Georgi and {llama.cpp contributors}},
  year         = {2023},
  howpublished = {\url{https://github.com/ggml-org/llama.cpp}},
}

@inproceedings{merity2017pointer,
  title     = {Pointer Sentinel Mixture Models},
  author    = {Merity, Stephen and Xiong, Caiming and Bradbury, James and Socher, Richard},
  booktitle = {International Conference on Learning Representations (ICLR)},
  year      = {2017},
}

@misc{llama32,
  author = {{Meta}},
  title = {{Llama 3.2}: Model Card},
  year = {2024},
  howpublished = {\url{https://huggingface.co/meta-llama/Llama-3.2-1B-Instruct}},
}

@misc{qwen25three,
  author = {{Qwen Team}},
  title = {{Qwen2.5-3B-Instruct}: Model Card},
  year = {2024},
  howpublished = {\url{https://huggingface.co/Qwen/Qwen2.5-3B-Instruct}},
}

@inproceedings{bfcl,
  title = {The Berkeley Function Calling Leaderboard ({BFCL}): From Tool Use to Agentic Evaluation of Large Language Models},
  author = {Patil, Shishir G and Mao, Huanzhi and Yan, Fanjia and Ji, Charlie Cheng-Jie and Suresh, Vishnu and Stoica, Ion and Gonzalez, Joseph E.},
  booktitle = {Proceedings of the 42nd International Conference on Machine Learning},
  pages = {48371--48392},
  year = {2025},
  editor = {Singh, Aarti and Fazel, Maryam and Hsu, Daniel and Lacoste-Julien, Simon and Berkenkamp, Felix and Maharaj, Tegan and Wagstaff, Kiri and Zhu, Jerry},
  volume = {267},
  series = {Proceedings of Machine Learning Research},
  month = {13--19 Jul},
  publisher = {PMLR},
  url = {https://proceedings.mlr.press/v267/patil25a.html}
}

@misc{morphserve,
  title = {{MorphServe}: Efficient and Workload-Aware {LLM} Serving via Runtime Quantized Layer Swapping and {KV} Cache Resizing},
  author = {Zhaoyuan Su and Zeyu Zhang and Tingfeng Lan and Zirui Wang and Haiying Shen and Juncheng Yang and Yue Cheng},
  year = {2026},
  eprint = {2506.02006},
  archivePrefix = {arXiv},
  primaryClass = {cs.DC},
  url = {https://arxiv.org/abs/2506.02006},
}

@inproceedings{qaq_query,
  title = {{QAQ}: Query-adaptive Mixed-precision Quantization for Large Language Models},
  author = {Li, Shuxing and Liu, Huanrong and Wang, Zelin and Du, Ruoyang and Lee, S H and Tian, Chunlin and Li, Qingbiao},
  booktitle = {ML for Systems Workshop at NeurIPS},
  year = {2025},
  url = {https://mlforsystems.org/assets/papers/neurips2025/paper68.pdf},
}

@misc{accuracy_not_all,
  title = {Accuracy is Not All You Need},
  author = {Abhinav Dutta and Sanjeev Krishnan and Nipun Kwatra and Ramachandran Ramjee},
  year = {2024},
  eprint = {2407.09141},
  archivePrefix = {arXiv},
  primaryClass = {cs.LG},
  url = {https://arxiv.org/abs/2407.09141},
}

@misc{kl_lens,
  title = {A {KL} Lens on Quantization: Fast, Forward-Only Sensitivity for Mixed-Precision {SSM-Transformer} Models},
  author = {Jason Kong and Nilesh Prasad Pandey and Flavio Ponzina and Tajana Rosing},
  year = {2026},
  eprint = {2604.13440},
  archivePrefix = {arXiv},
  primaryClass = {cs.LG},
  url = {https://arxiv.org/abs/2604.13440},
}

@article{williams2009roofline,
  author = {Williams, Samuel and Waterman, Andrew and Patterson, David},
  title = {Roofline: an insightful visual performance model for multicore architectures},
  year = {2009},
  issue_date = {April 2009},
  publisher = {Association for Computing Machinery},
  address = {New York, NY, USA},
  volume = {52},
  number = {4},
  issn = {0001-0782},
  url = {https://doi.org/10.1145/1498765.1498785},
  doi = {10.1145/1498765.1498785},
  journal = {Commun. ACM},
  month = apr,
  pages = {65--76},
  numpages = {12}
}

@inproceedings{nestedfp,
  title = {Nested{FP}: High-Performance, Memory-Efficient Dual-Precision Floating Point Support for {LLMs}},
  author = {Lee, Haeun and Kwon, Omin and Park, Yeonhong and Lee, Jae W.},
  year = {2025},
  eprint = {2506.02024},
  archivePrefix = {arXiv},
  booktitle = {Advances in Neural Information Processing Systems (NeurIPS)}
}

@inproceedings{anyprecision,
  title = {Any-Precision {LLM}: Low-Cost Deployment of Multiple, Different-Sized {LLMs}},
  author = {Park, Yeonhong and Hyun, Jake and Cho, SangLyul and Sim, Bonggeun and Lee, Jae W.},
  year = {2024},
  eprint = {2402.10517},
  archivePrefix = {arXiv},
  booktitle = {Proceedings of the 41st International Conference on Machine Learning (ICML)}
}

@inproceedings{matquant,
  title = {Matryoshka Quantization},
  author = {Nair, Pranav Ajit and Datta, Puranjay and Dean, Jeff and Jain, Prateek and Kusupati, Aditya},
  year = {2025},
  eprint = {2502.06786},
  archivePrefix = {arXiv},
  booktitle = {Proceedings of the 42nd International Conference on Machine Learning (ICML)}
}

@inproceedings{bitstack,
  title = {BitStack: Any-Size Compression of Large Language Models in Variable Memory Environments},
  author = {Wang, Xinghao and Wang, Pengyu and Wang, Bo and Zhang, Dong and Zhou, Yunhua and Qiu, Xipeng},
  year = {2025},
  eprint = {2410.23918},
  archivePrefix = {arXiv},
  booktitle = {International Conference on Learning Representations (ICLR)}
}

@misc{rrq,
  title = {Recurrent Residual Quantization: A Progressive Multi-Precision Representation for {LLMs}},
  author = {Luo, Yu and Dong, Bo and Cheng, Wenhua and Shen, Haihao},
  year = {2026},
  eprint = {2608.04048},
  url = {https://arxiv.org/abs/2608.04048},
  archivePrefix = {arXiv}
}

@misc{flexquant,
  title = {FlexQuant: A Flexible and Efficient Dynamic Precision Switching Framework for {LLM} Quantization},
  author = {Liu, Fangxin and Wang, Zongwu and Xia, JinHong and Zhao, Junping and Zhao, Shouren and Li, Jinjin and Liu, Jian and Jiang, Li and Guan, Haibing},
  year = {2025},
  eprint = {2506.12024},
  url = {https://arxiv.org/abs/2506.12024},
  archivePrefix = {arXiv}
}

@misc{dpllm,
  title = {{DP-LLM}: Runtime Model Adaptation with Dynamic Layer-wise Precision Assignment},
  author = {Kwon, Sangwoo and Seo, Seong Hoon and Lee, Jae W. and Park, Yeonhong},
  year = {2025},
  eprint = {2508.06041},
  url = {https://arxiv.org/abs/2508.06041},
  archivePrefix = {arXiv}
}

\clearpage
\appendix
\let\arxivoldsection\section
\renewcommand{\section}{\FloatBarrier\arxivoldsection}
\renewcommand{\thesection}{A\arabic{section}}
\counterwithin{figure}{section}
\counterwithin{table}{section}
\renewcommand{\thefigure}{\thesection.\arabic{figure}}
\renewcommand{\thetable}{\thesection.\arabic{table}}
\renewcommand{\theequation}{A\arabic{equation}}

\section{Allocator details}
\label{app:opt}
This section gives the accuracy-budget objective, the allocator's inputs, and how well its additive accuracy cost
tracks the error of a whole configuration.

For an accuracy budget $\epsilon$, allocation minimizes the predicted time change:
\begin{equation}
\min_{\{p_u\}}\ \sum_u\Delta T(u,p_u)\quad\text{s.t.}\quad
\sum_u\Delta A(u,p_u)\le\epsilon,\qquad p_u\in\mathcal{P}_u,
\label{eq:accuracybudget}
\end{equation}
where $\epsilon$ is the allowed increase in additive accuracy cost. Figure~\ref{fig:arch} shows the allocator's
inputs and output. The dynamic program rounds
costs up to discrete budget levels and checks the chosen configuration against the unrounded budget.

\begin{figure}[H]
    \centering
    \resizebox{0.95\textwidth}{!}{%
    \begin{tikzpicture}[
        font=\footnotesize, >={Stealth[]}, line cap=round,
        cell/.style={draw, rounded corners=0.5pt, fill=blue!4, align=center, font=\scriptsize,
                     minimum width=30mm, minimum height=4.4mm, inner sep=0.8mm},
        chip/.style={draw, rounded corners=0.5pt, align=center, font=\scriptsize,
                     minimum width=22mm, minimum height=4.4mm, inner sep=0.8mm},
        live/.style={draw, rounded corners=0.5pt, fill=orange!14, align=center, font=\scriptsize,
                     minimum width=26mm, minimum height=4.4mm, inner sep=0.8mm},
        hw/.style={draw, rounded corners=0.5pt, fill=green!12, align=center, font=\scriptsize,
                   minimum width=24mm, minimum height=4.4mm, inner sep=0.8mm},
        grp/.style={draw, rounded corners, inner sep=1.3mm},
        alloc/.style={draw, rounded corners, fill=blue!16, align=center,
                      minimum width=22mm, minimum height=13mm, font=\small\bfseries},
        layer/.style={draw, align=center, minimum height=4.2mm, minimum width=22mm, font=\scriptsize},
        glab/.style={font=\scriptsize\bfseries},
    ]
        \colorlet{cf4}{red!20}\colorlet{cf6}{orange!35}\colorlet{cf8}{yellow!45}\colorlet{cbf}{blue!35}
        \colorlet{loopc}{orange!80!black}
        \node[alloc] (alloc) {Dynamic\\program};
        \node[cell, left=13mm of alloc, yshift=19mm] (o1) {accuracy table $\Delta A$};
        \foreach \i/\p/\t in {2/o1/{reference latency $a,c$}, 3/o2/{group changes $\Delta T_g(p)$},
                              4/o3/{tensor weight counts $W(u)$}}
            \node[cell, below=0.7mm of \p] (o\i) {\t};
        \node[grp, fit=(o1)(o4), label={[glab]above:Offline measurements and inputs}] (off) {};
        \colorlet{cint}{purple!22}
        \node[chip, fill=cf8, below=10mm of o4] (s1) {Q8\_0};
        \foreach \i/\p/\c/\t in {2/s1/cf6/Q6\_K, 3/s2/cf4/Q4\_K, 4/s3/cbf/F16}
            \node[chip, fill=\c, below=0.7mm of \p] (s\i) {\t};
        \node[grp, fit=(s1)(s4), label={[glab]above:Supported precision menu}] (menu) {};
        \node[cell, above=9mm of alloc, minimum width=32mm] (p1) {target latency or accuracy};
        \node[grp, fit=(p1), label={[glab]above:User budget}] (priors) {};
        \node[layer, fill=cbf, right=17mm of alloc, yshift=9mm] (m1) {F16};
        \foreach \i/\p/\c/\t in {2/m1/cf6/Q6\_K, 3/m2/cf4/Q4\_K, 4/m3/cf8/Q8\_0}
            \node[layer, fill=\c, below=1mm of \p] (m\i) {\t};
        \node[layer, minimum height=3mm, below=1mm of m4] (md) {$\vdots$};
        \node[layer, fill=cf8, below=1mm of md] (mn) {Q8\_0};
        \node[grp, fit=(m1)(mn), label={[glab]above:Tensor configuration}] (model) {};
        \node[hw, right=10mm of model, minimum width=30mm, minimum height=8mm] (g1) {switch weights\\and decode};
        \node[grp, fit=(g1), label={[glab]above:Runtime}] (gemm) {};
        \node[live, below=13mm of alloc, minimum width=34mm, align=center] (rt)
            {context length $L$\\and measured cache error};
        \draw[->] (off.east) -- (alloc.west);
        \draw[->] (menu.east) -- (alloc.west);
        \draw[->] (priors.south) -- (alloc.north);
        \draw[->] (rt.north) -- (alloc.south);
        \draw[->, thick] (alloc.east) -- node[above, font=\scriptsize]{config} (model.west);
        \draw[->, thick] (model.east) -- node[above, font=\scriptsize]{run} (gemm.west);
        \coordinate (lp) at ([yshift=-7mm]rt.south);
        \draw[->, dashed, loopc, thick] (gemm.south) |- node[pos=0.75, below, font=\scriptsize]{as the KV cache grows, allocate for the next context range} (lp) -- (rt.south);
    \end{tikzpicture}}
    \caption{Allocator inputs and output.}

    \label{fig:arch}
\end{figure}

\subsection{How well additive KL tracks joint quality}
\label{sec:opdivergence}
\begin{figure}[H]
\centering
\includegraphics[width=\linewidth]{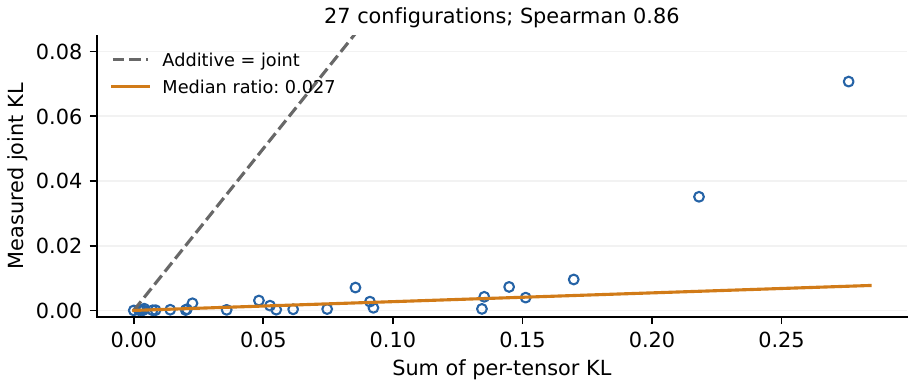}
\caption{Additive per-tensor accuracy cost versus measured joint KL for 27 fixed configurations. Gray shows equality and orange the median joint-to-additive ratio.}
\label{fig:calibration}
\end{figure}

The additive estimate ranks 27 Llama-3.2-1B-Instruct~\citep{llama32} configurations well (Spearman $0.86$ against
measured joint KL) but overestimates the joint KL itself (Fig.~\ref{fig:calibration}).

\section{Predictor calibration and evaluation}
\label{app:laptop_calibration}
This section gives how the predictor of Section~\ref{sec:predict} is fitted and its results for each held-out
configuration. The Q8 versus Q4 comparison in Figure~\ref{fig:latency_predictor} quantizes all 113 weight units of
Llama-3.2-1B-Instruct~\citep{llama32} on three 2048-token WikiText passages~\citep{merity2017pointer}.

\paragraph{Distributing group costs.}
To estimate each tensor's share of a measured group cost, the predictor uses the roofline work
estimate~\citep{williams2009roofline}
\begin{equation}
C(u,p)=\max\!\left(\frac{2W(u)M_u}{r_\mathrm{eff}(u,p)},\frac{W(u)b(p)}{\mathrm{BW}}\right).
\label{eq:predictor_work}
\end{equation}
Here $W(u)$ counts weights, $M_u$ counts token rows, $b(p)$ is bytes per weight including metadata,
$r_\mathrm{eff}$ is effective throughput, and $\mathrm{BW}$ is effective bandwidth.

The rate is $r_\mathrm{eff}(u,p)=(1-\eta_u)r(\mathrm{F16})+\eta_u r(p)$ with
$\eta_u=M_u/(M_u+M_{1/2})$, with values fitted from fixed-precision models ($\mathrm{BW}=67.6$\,GB/s,
$r(\mathrm{F16})=1.24$ and $r(\mathrm{Q8})=2.88$\,TFLOP/s, $M_{1/2}=0.07$ rows). These values only set proportions
within a group, and the measured group change sets its total:
\begin{equation}
\Delta T(u,\mathrm{F16})=
\frac{d_u}{\sum_{v\in g}d_v}\Delta T_g(\mathrm{F16}),\qquad
 d_u=C(u,\mathrm{F16})-C(u,\mathrm{Q8}).
\label{eq:predictor_group}
\end{equation}

\paragraph{Calibration and evaluation.}
Four controls (Q8 throughout, attention at F16, feed-forward at F16, and F16 throughout) give the reference time and
group changes (Table~\ref{tab:laptopfit}). The byte estimate fits an intercept and a weight-byte coefficient to the
same controls, and a constant estimate ignores precision.

\begin{figure}[H]\centering
\includegraphics[width=\linewidth]{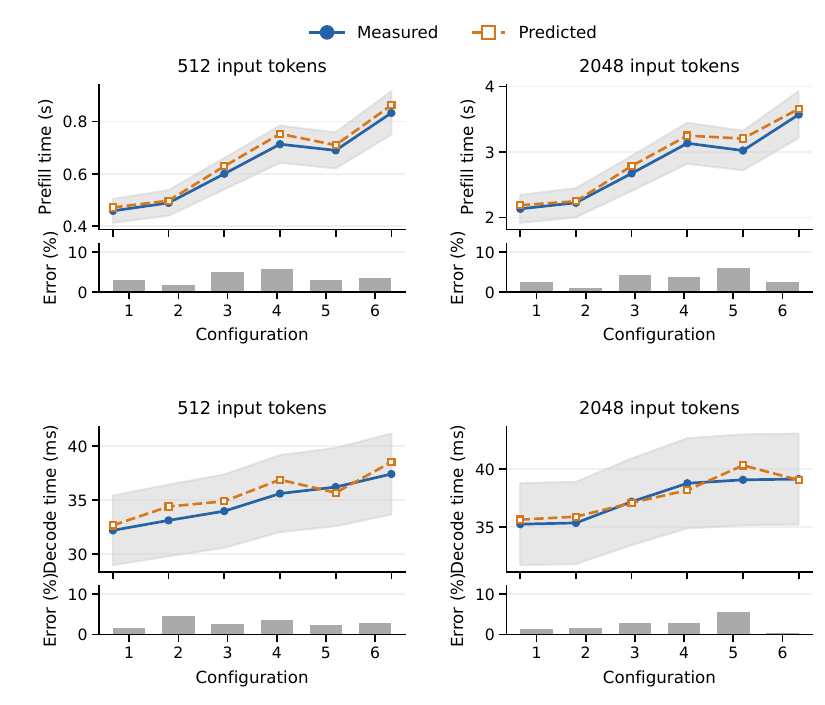}
\caption{Prediction at both calibrated prompt lengths, with configurations numbered as in
Figure~\ref{fig:latency_predictor}. Solid lines are measured, dashed lines predicted, and gray bands
mark $\pm10\%$ of measurement. Gray bars give mean absolute percentage error.}
\label{fig:laptop_predictor_detail}
\end{figure}

The six held-out configurations put F16 on key/value projections (1), query/output projections (2), random quarter,
half, and three-quarter subsets of units (3, 5, 6), and attention in the first half of layers with feed-forward in
the second half (4), with Q8 elsewhere (Figure~\ref{fig:laptop_predictor_detail}, Table~\ref{tab:predictor_errors}).

\begin{table}[H]\centering\small
\caption{Predictor calibration in ms. Columns give reference time $T_0$ and group changes
$\Delta T_g(\mathrm{F16})$ relative to Q8. We repeat calibration and evaluation in two separate runs, each calibrated
just before its evaluation.}
\label{tab:laptopfit}
\begin{tabular}{llrrrrr}\toprule Run & Phase & Context & $T_0$ & Attention & FFN & Output \\ \midrule
1 & prefill & 512 & 476.455 & 47.044 & 556.637 & -2.066 \\
1 & prefill & 2048 & 2211.277 & 110.744 & 2176.390 & 65.986 \\
1 & decode & 512 & 31.444 & 4.679 & 6.572 & -2.569 \\
1 & decode & 2048 & 33.461 & 1.620 & 6.536 & 1.109 \\
2 & prefill & 512 & 447.866 & 39.177 & 593.509 & -15.915 \\
2 & prefill & 2048 & 2124.196 & 92.931 & 2237.679 & 120.492 \\
2 & decode & 512 & 32.710 & 1.129 & 6.837 & 0.666 \\
2 & decode & 2048 & 37.648 & -0.770 & 3.082 & 3.873 \\
\bottomrule\end{tabular}\end{table}
\begin{table}[H]\centering\small
\caption{Mean absolute percentage error (\%) across all evaluation configurations and both prompt
lengths. All predictors use the same four calibration controls.}
\label{tab:predictor_errors}
\begin{tabular}{lrrr}\toprule Phase & Calibrated costs & Byte estimate & Constant estimate\\\midrule
Prefill & 3.51 & 5.92 & 17.62\\
Decode & 2.71 & 2.24 & 5.16\\
\bottomrule\end{tabular}
\end{table}

\section{Context-dependent schedules}
This section gives the calibration and full comparisons for the schedules of Section~\ref{sec:context}.

\subsection{Through 4K tokens}
\label{app:tensor_runtime}

Uniform F16/Q8 runs and runs that quantize one weight group at a time fit the timing model
(Table~\ref{tab:short_calibration}). At the last two scheduled changes, no configuration is predicted to meet the
target with the reserve, so we use the fastest predicted configuration.
\begin{table}[H]\centering\small
\caption{4K timing calibration, using contexts 256, 2048, and 4096. Group changes are relative to F16.}
\label{tab:short_calibration}
\begin{tabular}{llr}\toprule Quantity & Term & Value \\\midrule
Reference intercept (ms) & $a$ & 46.94 \\
Context growth (ms per 1024 tokens) & $c$ & 1.738 \\
Attention change at Q8 (ms) & $\Delta T_\mathrm{attention}(\mathrm{Q8})$ & $-1.26$ \\
Feed-forward change at Q8 (ms) & $\Delta T_\mathrm{FFN}(\mathrm{Q8})$ & $-6.29$ \\
Output change at Q8 (ms) & $\Delta T_\mathrm{output}(\mathrm{Q8})$ & $+0.66$ \\
Total reserve, including switching (ms) & & 4.06 \\
Switching allowance (ms) & & 2.50 \\
\bottomrule\end{tabular}
\end{table}

\begin{figure}[!tbp]
\centering
\includegraphics[width=0.78\linewidth]{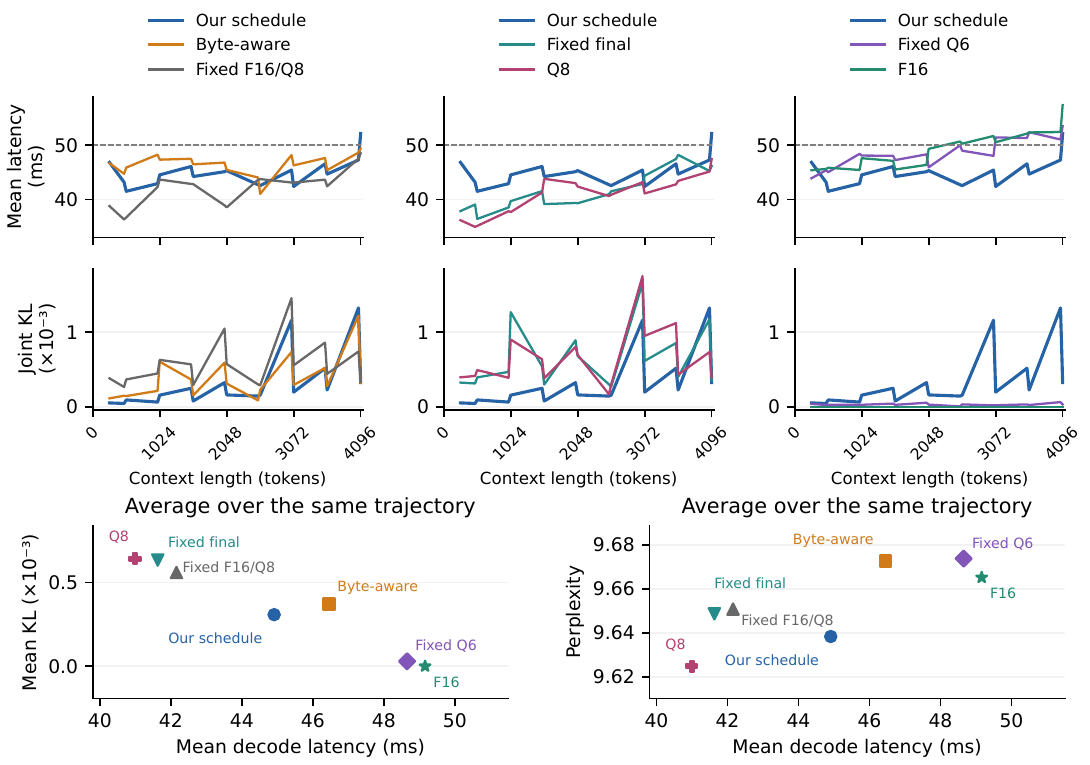}
\caption{All seven policies at the 17 contexts, with 50\,ms dashed. Bottom: mean latency versus mean joint KL to F16 or perplexity over the same passage.}
\label{fig:ctxall}
\end{figure}

Figure~\ref{fig:ctxall} adds all fixed configurations and the byte-aware schedule to the main comparison.

The extra weight copies add 1.61\,GB. Table~\ref{tab:switch_replay} repeats the larger tensor changes to measure
switching time.
\begin{table}[H]\centering\small
\caption{Additional switching check across three program runs. Times are ms. Median switch summarizes
repetitions after the first in each run. The last two columns give maximum first-token latencies,
including switching, for the first and later repetitions.}
\label{tab:switch_replay}
\begin{tabular}{rrrr}\toprule Tensors changed & Median switch (ms) & First token: initial (ms) & First token: later (ms)\\\midrule
27 & 1.68 & 36.58 & 22.50\\
44 & 1.60 & 23.15 & 22.19\\
\bottomrule\end{tabular}
\end{table}

\subsection{Through 16K tokens}
\label{app:long_protocol}

Each calibration configuration is timed between Q8 runs from the same saved state, which reduces the effect of
laptop speed drift (Table~\ref{tab:long_calibration}).
\begin{table}[H]\centering\small
\caption{16K calibration. Weight-group time changes $\Delta T_g(p)$ are relative to Q8, in ms.
The reference curve is $T_0(L)=a+cL/1024$.}
\label{tab:long_calibration}
\begin{tabular}{lrr}\toprule Group & Q6 & Q4\\\midrule
Attention & $-1.485$ & $-3.350$\\
Feed-forward & $-4.462$ & $-12.470$\\
Embedding/output & $-3.306$ & $-2.462$\\\midrule
\multicolumn{2}{l}{Reference intercept $a$ (ms)} & 41.474\\
\multicolumn{2}{l}{Context growth $c$ (ms per 1024 tokens)} & 3.045\\
\multicolumn{2}{l}{Prediction-error allowance (ms)} & 4.384\\
\multicolumn{2}{l}{Switching allowance (ms)} & 2.50\\
\multicolumn{2}{l}{Group-fit root mean squared error (ms)} & 1.87\\
\bottomrule\end{tabular}
\end{table}

\begin{figure}[H]\centering
\includegraphics[width=0.8\linewidth]{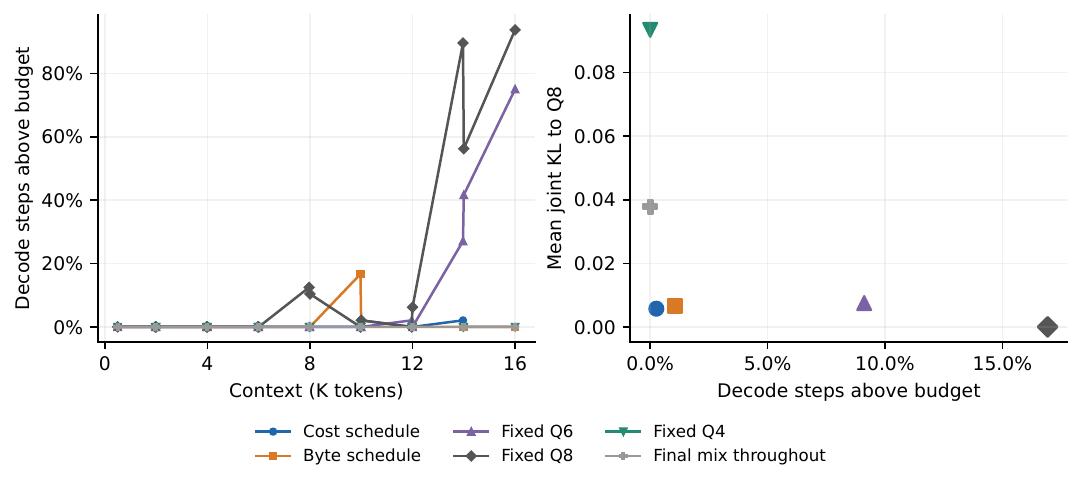}
\caption{Left: the percentage of decode steps exceeding 77\,ms at each context. Right: mean joint KL versus the overall percentage exceeding that target.}
\label{fig:long_attainment}
\end{figure}

Table~\ref{tab:long_results} and Figure~\ref{fig:long_attainment} give the full comparison on two held-out passages.
\begin{table}[H]\centering\small
\caption{16K results for all policies. Latencies are ms; P99 is the 99th-percentile decode latency.
The final column gives the percentage of steps within 77\,ms.}
\label{tab:long_results}
\begin{tabular}{lrrrrr}\toprule Policy & Mean (ms) & P99 (ms) & Joint KL & Perplexity & Within target (\%)\\\midrule
Our schedule & 58.12 & 72.86 & 0.00574 & 10.398 & 99.7\\
Byte-aware & 53.81 & 76.35 & 0.00662 & 10.447 & 99.0\\
Fixed final mix & 45.32 & 70.66 & 0.03785 & 10.946 & 100.0\\
Fixed Q8 & 62.16 & 85.18 & 0.00000 & 10.365 & 83.1\\
Fixed Q6 & 55.58 & 82.56 & 0.00753 & 10.436 & 90.9\\
Fixed Q4 & 39.49 & 64.50 & 0.09345 & 11.814 & 100.0\\
\bottomrule\end{tabular}\end{table}

The extra weight formats add 2.07\,GB, and a switch takes a median 1.55\,ms.

\section{Accuracy-budget schedule with a quantized cache}
\label{app:accuracy_recovery}\label{app:kvdegrad}
This section gives the calibration and full results for the accuracy-budget schedule of
Section~\ref{sec:accuracy_context}. The cache stays Q4\_0 throughout, and the quality reference uses Q8 weights with
an F16 cache.

\paragraph{Calibrating the accuracy budget.}
Table~\ref{tab:cache_calibration} gives the measured cache error, the budget it leaves for the weights, and the
resulting schedule. If the cache alone exceeds the budget, both allocators fall back to Q8 weights.
\begin{table}[H]\centering\small
\caption{Cache-error calibration and the resulting schedule. Weight shares count parameters.}
\label{tab:cache_calibration}
\begin{tabular}{rrrrrr}\toprule Context & Cache KL & Weight KL budget & Q8 (\%) & Q6 (\%) & Q4 (\%)\\\midrule
512 & 0.0325 & 0.0175 & 1.4 & 35.6 & 63.0\\
2048 & 0.0308 & 0.0192 & 1.0 & 29.1 & 69.9\\
4096 & 0.0755 & -0.0255 & 100.0 & 0.0 & 0.0\\
8192 & 0.1355 & -0.0855 & 100.0 & 0.0 & 0.0\\
12288 & 0.0366 & 0.0134 & 3.0 & 49.4 & 47.6\\
16384 & 0.0361 & 0.0139 & 3.0 & 46.5 & 50.5\\
\bottomrule\end{tabular}
\end{table}

The cache error need not grow smoothly with context, since it is measured on particular text. At 8K, one token
contributes 75\% of it.

\paragraph{Calibrating time.}
Table~\ref{tab:accuracy_time} gives the fitted time changes with the quantized cache.
\begin{table}[H]\centering\small
\caption{Time changes $\Delta T_g(p)$ with the Q4 cache, relative to Q8 weights, in ms.}
\label{tab:accuracy_time}
\begin{tabular}{lrr}\toprule Group & Q6 & Q4\\\midrule
Attention weights & $-0.147$ & $-4.296$\\
Feed-forward weights & $-6.986$ & $-11.610$\\
Embedding/output & $+0.265$ & $-3.151$\\
\bottomrule\end{tabular}
\end{table}

\begin{figure}[H]\centering
\includegraphics[width=\linewidth]{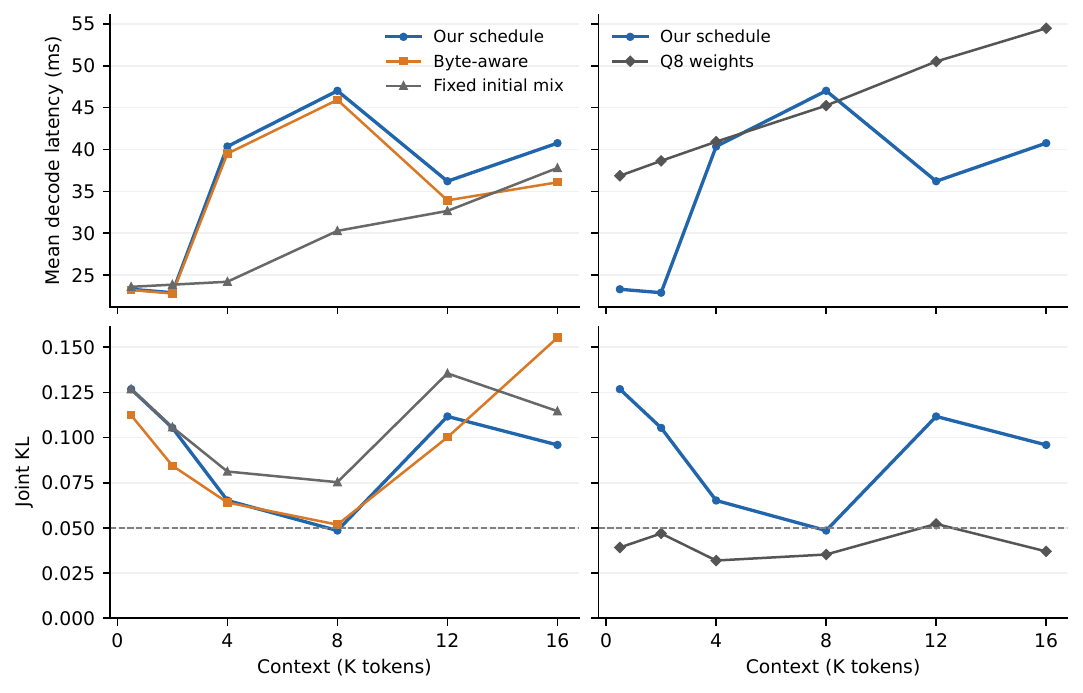}
\caption{Accuracy-budget comparisons at matching contexts. Left: our schedule and its fixed initial configuration. Right: our schedule and Q8 weights with the same Q4 cache. The dashed line marks the KL budget.}
\label{fig:accuracy_controls}
\end{figure}

\paragraph{Comparison.}
Table~\ref{tab:accuracy_results} and Figure~\ref{fig:accuracy_controls} add Q8 weights with the quantized cache,
which shows how much raising weight precision can recover while the cache stays quantized.
\begin{table}[H]\centering\small
\caption{Accuracy-budget results. The last column counts measurements with mean joint KL at most
0.05 across six context lengths in each of two passages. Mean latency is in ms. The Q8, F16-cache run is the quality reference and was not timed.}
\label{tab:accuracy_results}
\begin{tabular}{lrrrr}\toprule Policy & Mean (ms) & Joint KL & Perplexity & Within budget\\\midrule
Our schedule & 35.10 & 0.0923 & 12.958 & 3/12\\
Byte-aware & 33.59 & 0.0947 & 13.297 & 2/12\\
Fixed initial mix & 28.74 & 0.1065 & 12.896 & 1/12\\
Q8, Q4 cache & 44.45 & 0.0404 & 12.025 & 9/12\\
Q8, F16 cache & -- & 0.0000 & 11.199 & 12/12\\
\bottomrule\end{tabular}
\end{table}

\section{Tensor scheduling on tool-use turns}
\label{app:tensor_agents}
This section gives the calibration and results at every budget for the tool-use evaluation of
Section~\ref{sec:agents}.

\paragraph{Calibration and execution.}
We allocate six-layer groups plus embedding and output groups (Table~\ref{tab:agent_calibration}).
\begin{table}[H]\centering\small
\caption{Tool-use timing calibration. Group changes $\Delta T_g(\mathrm{Q4})$ are relative to Q8.
The reference curve is $T_0(L)=a+cL/1024$. Configurations are measured at contexts 512/4096,
with uniform Q8/Q4 also measured at 8192; each run has three warmup and 24 measured steps.}
\label{tab:agent_calibration}
\begin{tabular}{lr}\toprule Quantity & Value\\\midrule
Reference intercept $a$ (ms) & 78.036\\
Context growth $c$ (ms per 1024 tokens) & 2.751\\
Layers 1--6: time change (ms) & $-4.613$\\
Layers 7--12: time change (ms) & $-5.606$\\
Layers 13--18: time change (ms) & $-4.974$\\
Layers 19--24: time change (ms) & $-5.050$\\
Layers 25--30: time change (ms) & $-3.586$\\
Layers 31--36: time change (ms) & $-7.098$\\
Embedding: time change (ms) & $-4.518$\\
Output: time change (ms) & $+0.614$\\
Scheduling margin (ms) & 5.977\\
\bottomrule\end{tabular}
\end{table}

Schedules are chosen every 256 tokens, and each turn allows six model calls of up to 192 tokens.

\begin{figure}[H]\centering
\includegraphics[width=0.85\linewidth]{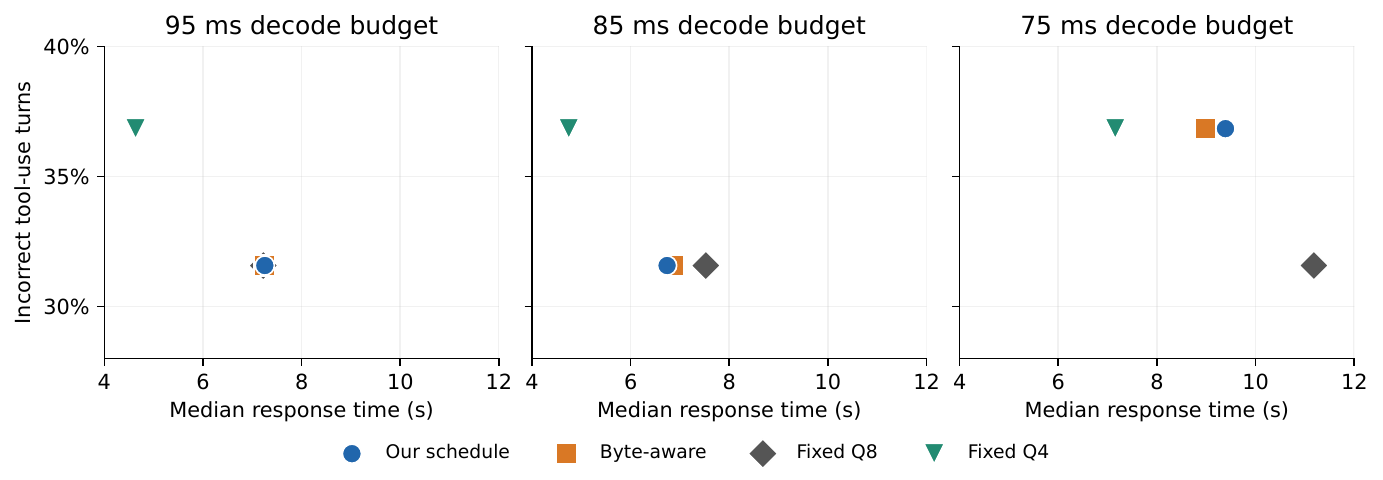}
\caption{Tool-use error versus median response time under each decode budget, over the same 19 turns.}
\label{fig:agent_all_budgets}
\end{figure}

\paragraph{Comparison across budgets.}
Figure~\ref{fig:agent_all_budgets} and Table~\ref{tab:agent_turns} give every budget. At the largest budget the
schedules mostly keep Q8. The smallest-budget run is slower even for fixed configurations, so its times also reflect
changed runtime conditions.
\begin{table}[H]\centering\small
\caption{Percentage of generated tokens decoded within each target. Median response times are in
Figure~\ref{fig:agent_all_budgets}.}
\label{tab:agent_turns}\label{tab:agent_decode}
\begin{tabular}{rlr}\toprule Decode target (ms) & Policy & Tokens within target (\%)\\\midrule
95 & \textbf{Our schedule} & \textbf{99.96}\\
95 & \textbf{Byte-aware} & \textbf{99.37}\\
95 & Fixed Q8 & 99.84\\
95 & Fixed Q4 & 100.00\\\midrule
85 & \textbf{Our schedule} & \textbf{99.20}\\
85 & \textbf{Byte-aware} & \textbf{99.44}\\
85 & Fixed Q8 & 59.84\\
85 & Fixed Q4 & 100.00\\\midrule
75 & \textbf{Our schedule} & \textbf{0.00}\\
75 & \textbf{Byte-aware} & \textbf{0.00}\\
75 & Fixed Q8 & 0.00\\
75 & Fixed Q4 & 74.83\\
\bottomrule\end{tabular}
\end{table}

\section{Executing schedules without resident copies}
\label{app:memory}
This section gives the runtime details and full measurements for Section~\ref{sec:memory}.

\paragraph{Runtime.} These measurements use a rebuilt llama.cpp runtime that gives each weight tensor its own buffer,
so versions can be freed one at a time. On it, a 40\,ms target gives a 16K schedule that keeps Q8 through 10K tokens
and then lowers precision in three steps (Figure~\ref{fig:stream_memory}).

\paragraph{Three executions.} Streamed versions are stored on disk as the runtime's own quantized bytes, so they match
the resident versions exactly. A reader thread loads the next interval's tensors, the decode thread uploads at most
4\,MB per step, and a third thread frees versions no later interval uses. Quantizing in place uses four CPU threads.
For residual branches, llama.cpp's graph builder computes $\hat W x+Rx$ for weights with a registered residual.

\paragraph{Measurement.} Laptop speed drifted during these measurements, so step times are not comparable across
runs (Table~\ref{tab:memory_runs}). To account for this, we compare each step that loads tensors with nearby steps in
the same run that do not.

\begin{table}[H]\centering\small
\caption{All runs, one process each.}
\label{tab:memory_runs}
\setlength{\tabcolsep}{4pt}
\begin{tabular}{llrrrrr}\toprule
Execution & Passage & Mean (ms) & Peak (GB) & At 16K (GB) & Loading time (s) & Perplexity\\\midrule
Resident & 0 & 48.7 & 3.23 & 3.23 & -- & 12.542\\
Resident & 1 & 51.5 & 3.23 & 3.23 & -- & 10.130\\
Streamed & 0 & 55.1 & 2.35 & 0.93 & 1.14 & 12.542\\
Streamed & 1 & 66.6 & 2.35 & 0.93 & 1.29 & 10.130\\
Quantized in place & 0 & 66.6 & 2.35 & 0.93 & 11.42 & 12.566\\
Quantized in place & 1 & 63.3 & 2.35 & 0.93 & 10.36 & 10.120\\
\bottomrule\end{tabular}\end{table}

\paragraph{Residual accuracy and speed.} Table~\ref{tab:residual} compares residual designs with Q8 and Q4\_K. Each
block is a separate session with its own Q8 baseline, so compare within a block. ``Stacked'' stores base and residual
as one tensor with twice the rows.

\begin{table}[H]\centering\small
\caption{Residual designs. KL is to F16 over 1{,}024 tokens; decode is the mean over 32 steps per configuration and
context, three rounds, in ms; prefill is one 512-token chunk. The three timed blocks are separate sessions.}
\label{tab:residual}
\setlength{\tabcolsep}{3.5pt}
\begin{tabular}{lrrrrrrr}\toprule
& & & \multicolumn{4}{c}{Decode at context} & Prefill\\\cmidrule(lr){4-7}
Weights & GB & KL & 512 & 4K & 8K & 16K & 512 tok.\\\midrule
Q8 & 1.59 & 0.00052 & 24.5 & 29.7 & 34.0 & 46.3 & 740\\
Q4\_K & 0.84 & 0.089 & 17.6 & 23.2 & 25.7 & 39.2 & 783\\
Stacked Q4\_K, residual off & 1.67 & 0.086 & 19.1 & 21.7 & 24.9 & 39.6 & 772\\
Stacked Q4\_K, residual on & 1.67 & 0.00082 & 35.5 & 38.3 & 40.3 & 54.5 & 1499\\\midrule
Q8 & 1.59 & 0.00052 & 34.0 & 40.7 & 44.1 & 60.5 & 962\\
Stacked Q4\_K, residual off & 1.67 & 0.086 & 26.2 & 31.6 & 37.2 & 50.0 & 983\\
Stacked Q4\_K, residual on (two halves) & 1.67 & 0.00082 & 49.7 & 53.9 & 58.8 & 75.4 & 2232\\
Stacked Q4\_K, residual on (fused) & 1.67 & 0.00081 & 37.1 & 42.6 & 48.0 & 62.2 & 2283\\\midrule
Q8 & 1.59 & 0.00052 & 23.4 & 28.5 & 30.1 & 38.9 & 557\\
Q4\_K & 0.84 & 0.089 & 17.6 & 22.9 & 24.2 & 32.2 & 524\\
Q4\_0 pair, residual off & 1.67 & 0.139 & 14.4 & 20.1 & 22.4 & 30.2 & 453\\
Q4\_0 pair, residual on (two halves) & 1.67 & 0.00102 & 25.6 & 29.5 & 33.8 & 40.4 & 955\\
Q4\_0 pair, residual on (fused) & 1.67 & 0.00104 & 23.5 & 26.9 & 30.3 & 38.4 & 944\\\midrule
Q8 as F16 (accuracy only) & -- & 0.00051 & \multicolumn{5}{c}{}\\
Q6 & 1.23 & 0.0051 & \multicolumn{5}{c}{}\\
Q4\_K + one 2-bit stage (F16) & -- & 0.0066 & \multicolumn{5}{c}{}\\
Q4\_K + two 2-bit stages (F16) & -- & 0.00050 & \multicolumn{5}{c}{}\\
\bottomrule\end{tabular}\end{table}

Turning residuals on or off takes a few milliseconds, and the next decode step is slower while llama.cpp rebuilds
its graph.

\paragraph{Schedules that raise precision again.} The accuracy-budget schedule is re-derived on the rebuilt runtime
with a 0.05 budget. In the compaction session, each compaction keeps the first 2{,}048 cached positions. For residual
branches, the allocator chooses residual on or off for each tensor (Tables~\ref{tab:updown} and~\ref{tab:compaction}),
and Table~\ref{tab:mapped} instead maps Q8 and Q6 to residual-on and Q4 to residual-off.

\paragraph{Equal-latency comparison.} Both executions run on the same runtime, are calibrated in alternating rounds,
and share a 78\,ms target (Table~\ref{tab:eqlat}). Two earlier attempts were discarded, one because the laptop slowed
about 2$\times$ between calibrations and one because it ran out of memory.

\begin{table}[H]\centering\small
\caption{Equal-latency comparison on one runtime (compaction session, shared 78\,ms target).}
\label{tab:eqlat}
\begin{tabular}{llrrrrr}\toprule
Passage & Execution & Peak & Mean step & P99 step & Joint KL & Perplexity\\\midrule
0 & Resident copies & 2.46\,GB & 30.7\,ms & 44.4\,ms & 0.0017 & 10.740\\
0 & Residual, Q4\_K base & 1.67\,GB & 33.7\,ms & 46.4\,ms & 0.0060 & 10.747\\
1 & Resident copies & 2.46\,GB & 31.9\,ms & 46.7\,ms & 0.0028 & 11.639\\
1 & Residual, Q4\_K base & 1.67\,GB & 34.2\,ms & 46.6\,ms & 0.0070 & 11.682\\
\bottomrule\end{tabular}\end{table}

\begin{table}[H]\centering\small
\caption{The Q8/Q6/Q4 schedules executed with resident copies and with Q4\_K residual branches (residual on for Q8
and Q6, off for Q4): joint KL at the sampled steps (mean, median) and the number of steps where the residual
execution is closer to the reference.}
\label{tab:mapped}
\begin{tabular}{llrrrrr}\toprule
& & \multicolumn{2}{c}{Resident copies} & \multicolumn{2}{c}{Residual branches} & Residual closer\\
\cmidrule(lr){3-4}\cmidrule(lr){5-6}
Scenario & Passage & Mean & Median & Mean & Median & (steps)\\\midrule
Accuracy budget & 0 & 0.061 & 0.052 & 0.077 & 0.039 & 20/33\\
& 1 & 0.052 & 0.048 & 0.053 & 0.042 & 15/33\\
Compaction session & 0 & 0.0021 & 0 & 0.0028 & 0.0011 & 8/54\\
& 1 & 0.0026 & 0 & 0.0036 & 0.0013 & 7/54\\
\bottomrule\end{tabular}\end{table}

\end{document}